\documentclass[letterpaper]{article} 

\ifdefined\aaaianonymous
    \usepackage[submission]{aaai2026}
\else
    \usepackage{aaai2026}
\fi

\nocopyright

\usepackage{times}
\usepackage{helvet}
\usepackage{courier}
\usepackage[hyphens]{url}
\usepackage{graphicx}
\usepackage{natbib}
\usepackage{caption}
\usepackage{booktabs}
\usepackage{amsmath}

\newcommand{\systemname}{WAR}

\ifdefined
\fi

\ifdefined\aaaianonymous
    \title{\systemname{}: Workload-Aware Rollouts for Synchronous Agentic Reinforcement Learning}
\else
    \title{\systemname{}: Workload-Aware Rollouts for Synchronous Agentic Reinforcement Learning}
\fi

\ifdefined\aaaianonymous
\author{Anonymous Submission}
\affiliations{}
\else
\author{
    Ryan Xu,
    Atlas Zhao,
    David Bao,
    Frank Du
}
\affiliations{
    \{cutelemon6, 547588521, 892404627, 1093782566\}@qq.com
}
\fi

\begin{document}

\maketitle

\begin{abstract}
Long-horizon rollout generation has become the dominant 
systems bottleneck in agentic reinforcement learning (RL). As agents 
interact with environments over many turns, trajectories rapidly 
grow to tens of thousands of tokens, making synchronous RL training 
increasingly constrained by rollout. We propose \systemname{}, a 
workload-aware rollout system that substantially accelerates 
synchronous agentic RL by jointly optimizing decoding and scheduling.

\systemname{} is built on a key observation: the optimal rollout 
optimization strategy depends on runtime load:

(1) Under low load, \systemname{} enables model-free 
speculative decoding with SuffixDecoding, which reuses suffix 
patterns from previously completed trajectories as speculative 
drafts for future rollouts. Unlike model-based drafters, 
SuffixDecoding introduces no additional draft model and avoids GPU 
contention with rollout generation.

(2) Under high load, where saturated batched decoding leaves 
limited room for speculative speedup, \systemname{} shifts the 
optimization focus to cache-aware scheduling. A global scheduler 
places requests across rollout replicas based on cache locality, 
trajectory progress and server load, reducing redundant KV-cache 
recomputation and mitigating load imbalance.

By combining decoding-level suffix reuse with system-level rollout 
scheduling, \systemname{} delivers robust throughput improvements 
across workload regimes without changing the underlying RL algorithm. 
\systemname{} improves long-context agentic rollout throughput by
$1.4\times$ under low load and up to $1.6\times$ under high load. These results 
show that \systemname{} removes a major rollout bottleneck in 
synchronous agentic RL and provides a practical path toward scalable 
long-context agent training.
\end{abstract}


\section{Introduction}

Large language models (LLMs) have demonstrated remarkable capabilities across a wide range of domains, 
including coding~\cite{claudeopus47}, mathematics~\cite{yan2025surveymathematicalreasoningera} 
and agent~\cite{cheng2025agentr1trainingpowerfulllm}. Reinforcement learning (RL) has become an 
effective approach for eliciting and enhancing these capabilities~\cite{Guo_2025}. A typical RL 
training step consists of three stages: rollout generation, reward computation, and model update. 
Among them, rollout generation is increasingly becoming the dominant systems bottleneck for agentic RL.

This bottleneck is particularly severe in long-horizon agentic tasks. Unlike single-turn generation, 
agentic rollouts require the model to interact with external environments over multiple turns. As tasks 
become more complex, agents need more rounds of interaction to gather observations, execute tools, 
inspect feedback and refine their responses. As a result, the context length of a trajectory can 
quickly grow to tens of thousands of tokens. Prior work has observed that rollout generation can 
dominate the end-to-end RL training time, contributing a large fraction of the total runtime~\cite{qin2026seeronlinecontextlearning}. 
The problem is further amplified by the long-tail nature of agentic rollouts: while most requests 
finish early, a small number of trajectories require substantially more computation due to longer 
reasoning chains or more environment interactions. In synchronous RL training, these stragglers force 
other workers to wait, leading to poor resource utilization and reduced rollout throughput.

One way to alleviate this issue is asynchronous rollout, which overlaps rollout generation with 
training~\cite{fu2026areallargescaleasynchronousreinforcement}. While asynchronous execution can improve 
hardware utilization, it introduces new challenges. The most important one is policy staleness: trajectories 
may be generated using outdated policy parameters, making the collected data less consistent with the 
current model. Asynchronous rollout also requires more complex coordination among rollout workers, reward 
computation, and model updates. Therefore, improving the efficiency of synchronous rollout remains an 
important and practical direction.

Existing synchronous RL framework, such as veRL~\cite{Sheng_2025}, typically relies on simple request 
scheduling policies, such as least-request scheduling combined with sticky sessions. Such policies work 
reasonably well under light workloads, where GPU resources are not fully saturated and request placement 
has limited impact. However, under heavy workloads, many concurrent long-context requests compete for limited 
GPU memory and computation. In this regime, simple load-based scheduling becomes insufficient: request 
placement must also account for KV-cache locality, trajectory progress, and server load. Otherwise, 
trajectories may be repeatedly scheduled to suboptimal replicas, causing redundant KV-cache recomputation 
and exacerbating load imbalance.

Rollout generation is essentially batched LLM inference, and can potentially benefit from inference 
acceleration techniques. Speculative decoding is a widely used lossless acceleration method, where a drafter 
proposes multiple tokens and the target model verifies them in parallel~\cite{li2025eaglespeculativesamplingrequires}. 
The drafter can be model-based~\cite{cai2024medusasimplellminference} or model-free~\cite{fu2024breaksequentialdependencyllm}. 
However, directly applying model-based speculative decoding to RL rollouts is challenging. A model-based drafter 
may compete with the target model for GPU computation and memory, reducing overall rollout throughput. In addition, 
RL training continuously updates the target model. If the drafter is kept fixed, the mismatch between the drafter 
and the updated target model may reduce the acceptance length, wasting GPU computation during verification. 
If the drafter is updated together with the target model, the RL training framework must maintain and synchronize 
an additional model, introducing extra complexity and GPU overhead. Speculative decoding is most effective 
when GPU resources are underutilized, as it increases parallelism by verifying multiple draft tokens in a 
single forward pass. Under heavy workloads, however, standard batched decoding already saturates GPU resources, 
and the additional overhead of multi-token verification can outweigh the reduction in decoding steps, 
leading to degraded inference performance.

These observations suggest that rollout optimization should be workload-aware. Under low load, normal decoding 
often underutilizes the GPU, leaving room for speculative decoding to improve hardware utilization and reduce 
decoding steps. Under high load, however, batched decoding already saturates GPU resources, leaving limited room for 
speculative speedup. Instead, system-level scheduling becomes more important under heavy workloads. As many 
long-context trajectories concurrently compete for limited GPU memory and computation, poor request placement 
can cause redundant KV-cache recomputation, uneven server utilization, and long-tail stragglers.

Motivated by this insight, we propose \systemname{}, a workload-aware rollout system for synchronous agentic 
RL training. \systemname{} jointly optimizes decoding and scheduling, and adapts its optimization strategy according 
to runtime load. Under low load, \systemname{} enables model-free speculative decoding with 
SuffixDecoding~\cite{oliaro2025suffixdecodingextremespeculativedecoding}. SuffixDecoding reuses suffix patterns 
from previously generated trajectories as speculative drafts for future rollouts. Since it is model-free, it 
introduces no additional draft model and avoids GPU contention with rollout generation. This makes it particularly 
suitable for agentic rollouts, where recurring code patterns may appear both across different prompts and among 
multiple rollouts of the same prompt.

Under high load, \systemname{} shifts the optimization focus to cache-aware scheduling. A global scheduler assigns 
requests to rollout replicas based on cache locality, trajectory progress, and server load. By favoring placements 
with higher KV-cache reuse and by prioritizing shorter trajectories when appropriate, the scheduler reduces redundant 
KV-cache recomputation, mitigates load imbalance, and preserves cache capacity for long-running trajectories. 

To keep the design simple and practical, \systemname{} uses runtime effective batch size as a lightweight workload 
indicator and determines the switching thresholds empirically for different workloads and hardware settings.

\systemname{} is designed to be lightweight and compatible with existing synchronous RL training pipelines. 
Built on top of veRL, it introduces a simple workload-aware control path that combines decoding-level suffix 
reuse with system-level rollout scheduling, without changing the underlying RL algorithm.

We make the following contributions:

\begin{itemize}
  \item \textbf{We identify workload as a first-class factor in rollout optimization.}
    We show that the effectiveness of rollout optimization techniques is workload-dependent: an optimization 
    that improves throughput under one workload regime may provide limited benefit, or even introduce overhead, 
    under another.

  \item \textbf{We design \systemname{}, a workload-aware rollout system for synchronous agentic RL.}
    \systemname{} combines model-free SuffixDecoding with cache-aware scheduling, enabling decoding-level reuse 
    under low load and reducing KV-cache recomputation and load imbalance under high load.

  \item \textbf{We demonstrate substantial rollout throughput improvements.}
    We implement \systemname{} on top of veRL with approximately 7K lines of Python code and evaluate it using 
    Qwen3-32B on production agent task datasets with 16 H100 GPUs. Compared with the veRL baseline, \systemname{} 
    improves long-context agentic rollout throughput by $1.4\times$ under low load and up to $1.6\times$ under high load.
\end{itemize}
\section{Background}
\label{sec:background}

\subsection{Synchronous Agentic RL Rollout}
\label{sec:bg_sync_rollout}

Reinforcement learning (RL) has become an important post-training technique for improving the reasoning and 
problem-solving capabilities of large language models. A typical RL iteration consists of three major 
stages~\cite{shao2024deepseekmathpushinglimitsmathematical}: rollout, reward computation, and training. 
During rollout, the actor model generates trajectories for a batch of prompts; during reward computation, 
the generated trajectories are evaluated by rule-based checkers, sandbox execution, reward models, 
or LLM-as-a-Judge~\cite{son2024llmasajudgerewardmodel}; finally, the actor model is updated using the collected 
training signals.

Many RL training systems perform rollout synchronously to preserve on-policy semantics. In this setting, 
the rollout stage must finish before the training stage starts, and all trajectories are generated by the 
most recent actor model. This design improves training stability and makes debugging and reproducibility easier. 
However, the synchronization barrier also exposes rollout inefficiency: when rollout lengths are imbalanced, 
workers that finish short trajectories early must wait for the slowest trajectories to complete. Prior systems 
have observed that rollout often dominates RL training time and becomes the primary source of GPU underutilization 
in synchronous RL post-training~\cite{gao2025rollpackermitigatinglongtailrollouts,qin2026seeronlinecontextlearning}.

Asynchronous rollout systems attempt to reduce such idle time by overlapping rollout and 
training~\cite{sheng2025laminarscalableasynchronousrl,fu2026areallargescaleasynchronousreinforcement,zhong2025streamrlscalableheterogeneouselastic,hu2026dorascalableasynchronousreinforcement}. 
However, asynchronous execution may introduce policy staleness, because trajectories generated by older model 
parameters can be used to update a newer policy. It can also introduce distributional skew, where faster-to-generate 
short samples disproportionately appear in earlier training batches. These effects may harm algorithmic fidelity 
and complicate debugging and reproducibility. Therefore, optimizing synchronous rollout remains an important 
problem for practical agentic RL training.

\subsection{Rollout Bottlenecks in Long-Horizon Agentic RL}
\label{sec:bg_bottleneck}

Agentic RL further amplifies the rollout bottleneck. Unlike single-turn generation, an agent interacts 
with an external environment over multiple turns. For example, in software engineering (SWE) tasks, an agent 
may inspect files, execute commands, observe tool outputs, and iteratively refine its solution. 
As the number of interaction turns increases, the context length of each trajectory grows rapidly, 
often reaching tens of thousands of tokens.

Long-context rollout introduces two major system challenges. First, the KV-cache footprint grows with the 
sequence length. A long-reasoning request may start with a small memory footprint but expand to tens of 
gigabytes as decoding progresses~\cite{qin2026seeronlinecontextlearning}. This dynamic memory growth can 
force batch-size shrinkage, trigger request preemption, and cause expensive re-prefill operations, reducing 
overall rollout throughput.

Second, rollout lengths are highly imbalanced. Long-tail response length is a key source of inefficiency 
in synchronous rollout. Since the next training stage cannot proceed until all trajectories in the current 
rollout step finish, a small number of long trajectories can leave many rollout workers idle. This creates 
GPU bubbles and reduces end-to-end training efficiency. RollPacker reports that the longest response in a 
rollout batch can be tens of times longer than the median response, causing prolonged idle periods for GPUs 
assigned to shorter responses~\cite{gao2025rollpackermitigatinglongtailrollouts}.

These challenges suggest that rollout optimization must address both computation and memory efficiency. 
Simply increasing parallelism or adding more workers is insufficient: poor request placement can increase 
KV-cache recomputation, while long-tail trajectories can still leave many workers idle near the end of a rollout step.

\subsection{Why Workload Matters}
\label{sec:bg_workload}

Existing rollout optimizations are not uniformly effective across all runtime workloads. In particular, 
the bottleneck of rollout generation changes with the serving load.

Under low load, GPU resources are often underutilized. In this regime, decoding-level acceleration such as 
speculative decoding~\cite{leviathan2023fastinferencetransformersspeculative, chen2023acceleratinglargelanguagemodel} 
can improve throughput by increasing parallelism within each target-model forward pass. 
Speculative decoding proposes multiple draft tokens and verifies them with the target model in parallel while 
preserving the target output distribution. This property makes it attractive for RL rollout, where changing the 
trajectory distribution may affect training correctness.

However, under high load, standard batched decoding already saturates GPU resources, leaving limited room for 
speculative speedup. In this regime, speculative decoding still introduces additional overhead, such as 
draft construction, suffix-tree lookup, verification, and auxiliary attention-mask handling. 
When this overhead outweighs the benefit of accepted draft tokens, speculative decoding can degrade throughput 
rather than improve it.

Instead, system-level scheduling becomes more important under high load. Many long-context trajectories 
concurrently compete for limited GPU memory and computation. Inefficient request placement can reduce KV-cache 
locality, trigger redundant prefill computation, and worsen load imbalance across rollout replicas. 
Therefore, high-load rollout requires scheduling policies that account for KV-cache reuse, trajectory progress, 
and server load.

\subsection{Workload-Aware Rollout Optimization}
\label{sec:bg_workload_aware}

The above observations motivate a workload-aware view of rollout optimization. Rather than applying a single 
optimization uniformly, a rollout system should adapt its strategy to the current workload regime. Under low load, 
it should exploit underutilized GPU resources with decoding-level acceleration. Under high load, it should avoid 
unnecessary speculative overhead and instead focus on cache locality and load balancing.

\begin{figure}[t]
    \centering
    \includegraphics[width=0.9\linewidth]{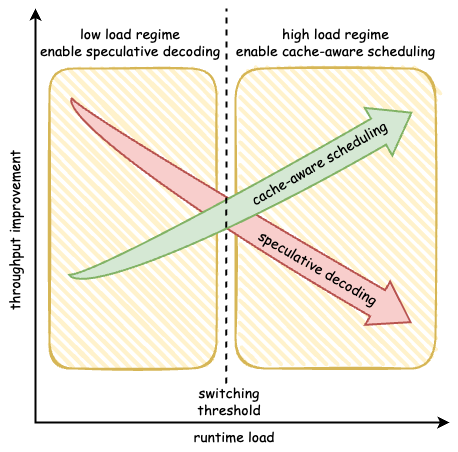}
    \caption{Motivation of WAR based on observation. In low load regime, multi-token verification of speculative decoding 
    can fully utilize underutilized gpu resources. In high load regime where GPU resources are fully utilized, 
    cache-aware scheduling is triggered to achieve a better performance by avoiding redundant prefill computation and load 
    imbalance. These two techniques are combined to achieve a overall better rollout performance.}
    \label{fig:WARDiagram}
\end{figure}

\systemname{} follows this principle, as shown in Figure~\ref{fig:WARDiagram}. Under low load, it enables model-free speculative decoding with 
SuffixDecoding~\cite{oliaro2025suffixdecodingextremespeculativedecoding}, which reuses suffix patterns from previously 
completed trajectories without introducing an additional draft model. Under high load, it shifts the optimization 
focus to cache-aware scheduling, reducing redundant KV-cache recomputation and mitigating load imbalance. 
By combining decoding-level suffix reuse with system-level scheduling, \systemname{} targets high rollout throughput 
across different workload regimes while preserving synchronous RL training semantics.

\section{\systemname{} System Design}

This section presents \systemname{}, a workload-aware rollout system for long-context agentic RL training. 
We start with an overview of the architecture and then describe the design of its main modules in detail. 

\subsection{Overview of \systemname{}}

\begin{figure}[t]
    \centering
    \includegraphics[width=1.0\linewidth]{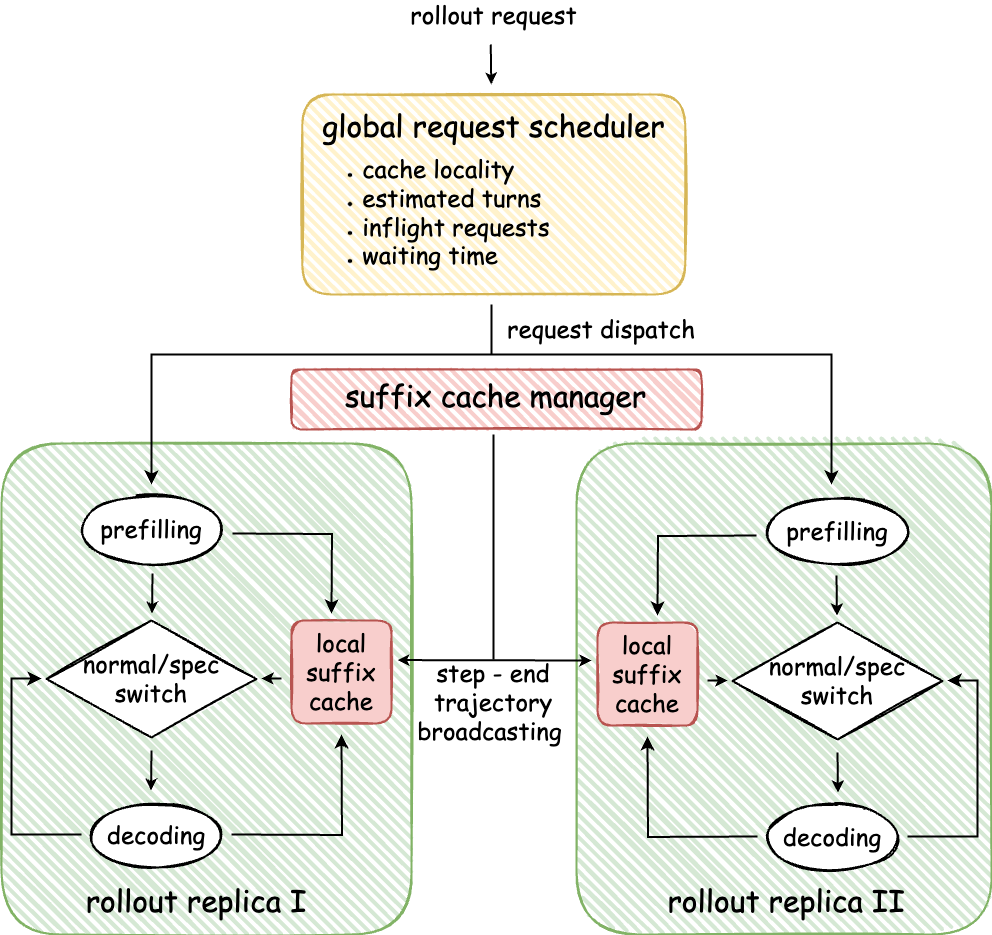}
    \caption{Architecture of WAR. WAR separates workload-aware optimization into two control levels. The global 
    scheduler performs cache-aware request placement across rollout replicas, while each inference instance 
    independently enables or disables SuffixDecoding according to its local effective batch size. At the end of 
    each RL step, newly generated trajectories are broadcast to all replicas to synchronize their local suffix caches. }
    \label{fig:WARArch}
\end{figure}

\systemname{} is centered around two key techniques: cache-aware scheduling and model-free speculative decoding, as shown in Figure~\ref{fig:WARArch}. 

In agentic RL, agents interact with environments over multiple turns, causing the context length of a trajectory 
to grow to the order of 10K--100K tokens. A rollout step is typically served by multiple rollout replicas, i.e., 
inference instances. If different turns of the same trajectory are frequently scheduled across different replicas, 
the system may repeatedly recompute KV caches, introducing non-negligible overhead. This overhead becomes especially 
significant for long-horizon multi-turn tasks such as SWE agent. Therefore, \systemname{} incorporates cache locality 
into rollout scheduling by considering the cache hit rate of each trajectory when assigning requests to rollout replicas.

\systemname{} further accelerates rollout generation with model-free speculative decoding. Speculative decoding is a widely 
used inference acceleration technique in which a verifier, usually the target model, verifies draft tokens generated by a 
drafter. The drafter can be either model-based or model-free. In RL rollouts, however, a model-based drafter may compete 
with the target model for GPU resources, reducing the overall rollout throughput. In contrast, SuffixDecoding is model-free 
and avoids such resource contention. SuffixDecoding stores the historical trajectories and matches the current token pattern 
against them to generate draft tokens. The workflow of GRPO-like algorithms is naturally well suited for SuffixDecoding. 
In these algorithms, each request is rolled out $n$ times to generate multiple trajectories. Since these trajectories 
originate from the same prompt and interact with similar task environments, they often exhibit repeated local string 
patterns. Based on this observation, \systemname{} integrates SuffixDecoding into the rollout pipeline to improve generation 
efficiency without changing the underlying RL algorithm.

\subsection{Cache-Aware Scheduling}
\label{sec:cacheAwareScheduling}

\begin{figure}[t]
    \centering
    \includegraphics[width=0.9\linewidth]{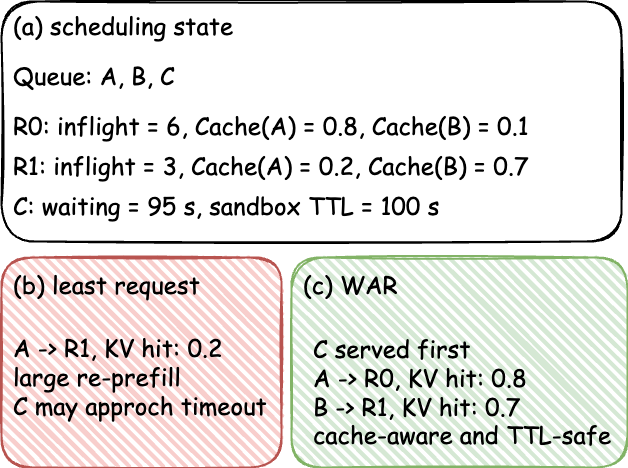}
    \caption{Illustrative example of cache-aware scheduling in WAR. The example shows a scheduling snapshot with 
    two rollout replicas and three queued requests. The least-request policy favors the less-loaded replica without 
    explicitly considering KV-cache locality. WAR first dispatches the request approaching its sandbox TTL, and then 
    places the remaining requests according to cache locality, estimated trajectory length, and replica load. Request A 
    is a new rollout whose sibling trajectories have populated the prefix cache on Replica 0, but it does not yet have 
    an established sticky-session placement. }
    \label{fig:WARSchedExample}
\end{figure}

At the begining of rollout, all rollout prompts are queued into a global scheduler. When the number of total 
request in the global request scheduler \(N_{\text{req,tot}}\) is less than a threshold \(Thr_{\text{sched}}\). The scheduling policy 
falls back to least-request combined with sticky-session, which is original scheduling policy of veRL. When 
\(N_{\text{req,tot}} \ge Thr_{\text{sched}}\), we score all the queued prompts 
to quantify which queued request should be scheduled. An illustrative example can be found in Figure~\ref{fig:WARSchedExample}. 
The priority is scored around several dimensions:

\begin{itemize}
    \item \textbf{Cache hit rate \(H\).}
    A higher \(H\) of a request means that current routed rollout instance has a higher kv cache in 
    its GPU global memory. Thus \(H\) is a locality affinity metric and can be directly obtained from 
    rollout server.

    \item \textbf{Online estimated assistant turns \(\hat{T}_g\).} 
    Beside cache hit rate, the actual length of context is also important. For two requests with same cache hit rate, 
    their actual length of kv cache may be different. It is observed that inside a rollout group, in which all trajectories 
    have same initial prompt, their assistant turns are similar. Assistant turn is usually proportional to context length. 
    Compared to exact context length, the assistant turns are easier to estimate. Once a multi-turn trajectory finishes 
    rollout, the corresponding assist turn can be updated as a function of group id:
    \[
    \hat{T}^{\,\text{new}}_g
    =
    \alpha T^{\text{obs}}_g
    +
    (1-\alpha)\hat{T}^{\,\text{old}}_g,
    \]
    where \(T_g^{\text{obs}}\) is the observed number of assistant turns for group \(g\) and \(\alpha\) is the smoothing 
    factor.

    \item \textbf{Number of inflight requests \(N_{\text{rep}}^{\text{inflight}}\) on each rollout replica \(\text{rep}\).} 
    It is a metric of workload on \(\text{rep}\).

    \item \textbf{Waiting time of each request \(t_{\text{req}}^{\text{waiting}}\) in the queue of global scheduler.}
    \(t_{\text{req}}^{\text{waiting}}\) is a practical concern because the sandbox related to the request have a prescribed 
    time to live \(\text{ttl}_{\text{sandox}}\). If the sandbox is not visited for \(\text{ttl}_{\text{sandox}}\), it will 
    be released. Therefore, a request can not stay to long in the queue.

\end{itemize}

Above 4 factors are combined together linearly to give a final score of a request:
\[
\begin{aligned}
S
&=
S_\text{waiting}
+
S_\text{turn}
+ 
S_\text{cache}
+
S_\text{inflight}
\\
&=
\mathbf{1}\!\left(t_{\text{req}}^{\text{wait}} > \text{scale}_\text{ttl} \cdot \text{ttl}_{\text{sandox}}\right)
\cdot \text{scale}_\text{waiting} \cdot t_{\text{req}}^{\text{wait}} 
\\
&\quad+
\text{scale}_\text{turn} \cdot \hat{T}_g
+
\text{scale}_\text{cache} \cdot H
+
\text{scale}_\text{inflight} \cdot N_{\text{rep}}^{\text{inflight}}.
\end{aligned}
\]
where
\[
S_\text{waiting} > 0,
\\
S_\text{turn} < 0,
\\
S_\text{cache} > 0,
\\
S_\text{inflight} < 0,
\\
0 < \text{scale}_\text{ttl} < 1.
\]
A negative \(\text{scale}_\text{turn}\) biases the scheduler toward shorter trajectories, enabling them to 
complete earlier and freeing KV cache for longer trajectories. Similarly, a negative \(\text{scale}_\text{inflight}\) 
biases scheduling toward less loaded servers, reducing load imbalance across rollout replicas. 
\(\text{scale}_\text{waiting}\), \(\text{scale}_\text{turn}\), \(\text{scale}_\text{cache}\), and 
\(\text{scale}_\text{inflight}\) are determined experimentally such that the score components have the 
following order of magnitude:
\[
|S_\text{waiting}|
\gg
|S_\text{turn}|
\gg
|S_\text{cache}|
\gg
|S_\text{inflight}|.
\]
Once the sandbox reaches \(\text{ttl}_{\text{sandbox}}\), the trajectory is terminated, resulting in an 
incomplete rollout. Therefore, \(|S_\text{waiting}|\) is assigned the largest magnitude to prioritize 
requests approaching the sandbox timeout. The coefficient \(\text{scale}_\text{ttl}\) is set between 0 
and 1 to account for additional overheads, such as scheduling and communication, ensuring that a request 
is dispatched before the sandbox shuts down. \(|S_\text{turn}|\) is the second largest because we want 
short trajectories finished as early as possible.

\(|S_\text{inflight}|\) is assigned the smallest magnitude because request load balancing is primarily 
handled by a stronger strategy inspired by CONCUR~\cite{chen2026concurhighthroughputagenticbatch}. CONCUR 
is a cache-aware request control algorithm based on the Additive Increase Multiplicative Decrease (AIMD) 
algorithm in TCP Congestion Control. Whether or not a rollout replica has free slot to a request is based 
on the KV cache Usage \(U\):
\[
W_{t} =
\begin{cases}
W_{t-1} + \alpha, & \text{if } U_{t-1} < U_\text{low},\\
W_{t-1} \times \beta, & \text{if } U_{t-1} > U_\text{high} \text{ and }  H_{t-1} < H_\text{thresh},\\
W_{t-1}, & \text{otherwise}.
\end{cases}
\]
where \(W_t\) and \(W_{t-1}\) are the maximum number of request on each inference instance in successive update time. 
We modify the algorithm as follows:
\[
\begin{aligned}
\hat{W}_{t} 
&=
\begin{cases}
W_{t - \delta} + \alpha, & \text{if } \bar{H} < H_\text{low},\\
W_{t - \delta} \times \beta, & \text{if } \bar{H} > H_\text{high},\\
W_{t - \delta}, & \text{otherwise}.
\end{cases}
\\[0.8em]
W_{t}
&=
\min\left(
\max\left(1, \hat{W}_{t}\right),
N^{\text{inflight}}_\text{rep} + \frac{N^{\text{queue}}_\text{req}}{N_\text{rep}}
\right).
\end{aligned}
\]
where \(H_\text{low}\) and \(H_\text{high}\) are the low and high threshold of cache hit rate. 
\(\bar{H}\) is the mean cache hit rate for the successive \(\delta\) requests. \(W_{t - \delta}\) 
is the maximum number of request before \(\delta\) requests, meaning that \(W_t\) is not updated 
until \(\delta\) requests are finished in order to stablize the maximum limit on each rollout replica. 
We bound \(W_t\) between 1 and \(N^{\text{inflight}}_\text{rep} + (N^{\text{queue}}_\text{req} / N_\text{rep})\), 
meaning that at least 1 request can be served on each rollout replica and the increased requests should 
not be more than number of queued requests averaged by number of rollout replica.

\subsection{SuffixDecoding}
SuffixDecoding~\cite{oliaro2025suffixdecodingextremespeculativedecoding} is a model-free speculative decoding 
algorithm based on suffix trees. It caches token-sequence patterns from previously generated trajectories and 
uses them during decoding to predict likely subsequent tokens. These predicted tokens are then verified by the 
target model, allowing the rollout process to advance by multiple tokens in a single forward pass when the 
drafts are accepted. SuffixDecoding is well suited for agentic RL rollouts for the following reasons:

\begin{itemize}
    \item It exploits token patterns from real generated trajectories, which can lead to high acceptance 
    rates in workloads with repeated local structures. 

    \item It introduces no additional draft model and therefore avoids GPU contention with rollout generation. 

    \item It incurs modest memory overhead, as it stores token-sequence structures rather than extra model parameters.
\end{itemize}

In agentic RL training, the rollout workload varies across steps and across servers. When the effective batch size 
is small, SuffixDecoding can provide significant acceleration because draft verification introduces limited overhead 
while accepted drafts allow the model to advance multiple tokens per step. However, when the effective batch size 
becomes large, the benefit of speculative decoding can diminish: the additional verification for draft tokens may 
offset the speedup from accepted tokens, leading to reduced throughput. This motivates an adaptive rollout system 
that applies SuffixDecoding when it is beneficial while avoiding unnecessary overhead under high-load conditions.

We implement SuffixDecoding in inference engine based on its open sourced code base with additional features:

\begin{itemize}
    \item \textbf{Adaptive enabling and disabling.} 
    We dynamically enable or disable speculative decoding according to the current batch size, allowing the system 
    to exploit suffix decoding under low-load conditions while avoiding unnecessary overhead under high load.

    \item \textbf{Stateless mode switching.} 
    We ensure safe switching between speculative and non-speculative execution without introducing state inconsistency 
    or memory leaks.

    \item \textbf{CUDA Graph compatibility.} 
    We support separate CUDA Graph capture and replay for speculative and non-speculative execution modes, enabling 
    efficient execution under both settings.

    \item \textbf{External update interface.} We expose APIs that allow external frameworks to proactively monitor 
    and update the suffix tree with newly generated trajectories, enabling cross-step and cross-worker suffix reuse.
\end{itemize}

In RL training, each rollout worker maintains its own suffix tree independently. When a request is dispatched to a 
worker, the generated trajectory is only used to update the local suffix tree of that worker. This design leads to 
several limitations:

\begin{itemize}
    \item Generated sequence patterns cannot be shared across different rollout workers.

    \item Responses from similar prompts cannot be effectively reused by other workers.

    \item Each worker can only leverage its own local history, resulting in limited historical coverage.
\end{itemize}

To address these limitations, we design a cross-worker suffix tree synchronization mechanism in the RL 
framework:
\begin{itemize}
    \item After each RL step, broadcast the complete prompt-response sequences generated in that step to all rollout workers.

    \item Use the HTTP interface of the Suffix Cache Server to asynchronously update the suffix trees.

    \item Avoid blocking the main training pipeline whenever possible.
\end{itemize}
\section{Implementation}
\systemname{} is implemented on top of veRL v0.8.0. We use SGLang v0.5.5 as the rollout inference engine and 
integrate SuffixDecoding into its inference pipeline. The reward and training components of veRL remain unchanged, 
with Megatron-LM v0.13.1 used as the training backend. To support software engineering tasks, we additionally 
implement an SWE agent in veRL, enabling multi-turn interaction with a customized sandbox environment.
\section{Evaluation}
\systemname{} is evaluated with Qwen3-32B on an SWE dataset containing 500 prompts. 
All experiments are conducted on an InfiniBand-connected H100 cluster with 2 nodes and 8 GPUs per node. 
Qwen3-32B is extended with YaRN~\cite{peng2026yarnefficientcontextwindow} to support a 96K context length.

The training-side tensor parallelism (TP), pipeline parallelism (PP), and context parallelism (CP) are 
configured as 4, 2, and 2, respectively, while the rollout-side TP is set to 4. 
These parallelism settings are kept unchanged throughout the evaluation. To evaluate \systemname{} across 
different workload regimes, we vary the training batch size from 4 to 64, specifically using batch sizes 
of 4, 8, 16, 32, and 64. Each prompt is rolled out 8 times, and each experiment is run for 5 training steps.

We use end-to-end rollout time, prefill throughput, and decode throughput as the primary metrics. 
We also report cache hit rate and acceptance length as auxiliary metrics to analyze the behavior of 
cache-aware scheduling and SuffixDecoding.

\begin{table}[t]
\centering
\caption{Workload differences relative to \texttt{veRL\_base}.
Values report the mean relative difference (\%) across training batch
sizes from 4 to 64. All differences remain within 5\%, indicating
comparable generated workloads across configurations.}
\label{tab:workload-comparability}
\begin{tabular}{lccc}
\toprule
& \textbf{sched\_only}
& \textbf{spec\_only}
& \textbf{sched\_spec} \\
\midrule
Prefill tokens   & $2 \pm 4$ & $1 \pm 2$ & $-2 \pm 1$ \\
Decode tokens    & $1 \pm 4$ & $0 \pm 4$ & $-2 \pm 2$ \\
Assistant turns  & $1 \pm 2$ & $0 \pm 2$ & $-1 \pm 1$ \\
\bottomrule
\end{tabular}
\end{table}

We set the temperature, \(\text{top}_p\), and \(\text{top}_k\) to 1.0, 1.0, and 200, respectively, 
to maintain the diversity of rollout trajectories. Data shuffling is disabled to facilitate fair comparison 
across different experiment settings. The maximum number of assistant turns is set to 120 to allow sufficient 
interaction between the agent and the sandbox.

To ensure that each experiment setting is comparable to the veRL baseline, we select only runs whose relative 
differences in prefill tokens, decode tokens, and assistant turns are all within 5\% compared with the veRL baseline, 
as shown in Table~\ref{tab:workload-comparability}. This filtering reduces the impact of workload variation and ensures 
that the measured throughput differences mainly reflect system-level optimizations rather than differences in 
generated workload.

\subsection{End-to-End Evaluation}

To understand the contribution of each component, we evaluate four configurations that enable or disable 
cache-aware scheduling and SuffixDecoding independently:

\begin{itemize}
    \item \textbf{veRL\_base.} The baseline configuration where both cache-aware scheduling and SuffixDecoding are disabled.
    \item \textbf{sched\_only.} Only cache-aware scheduling is enabled.
    \item \textbf{spec\_only.} Only SuffixDecoding is enabled.
    \item \textbf{sched\_spec.} Both cache-aware scheduling and SuffixDecoding are enabled.
\end{itemize}

\begin{figure}[t]
    \centering
    \includegraphics[width=1.0\linewidth]{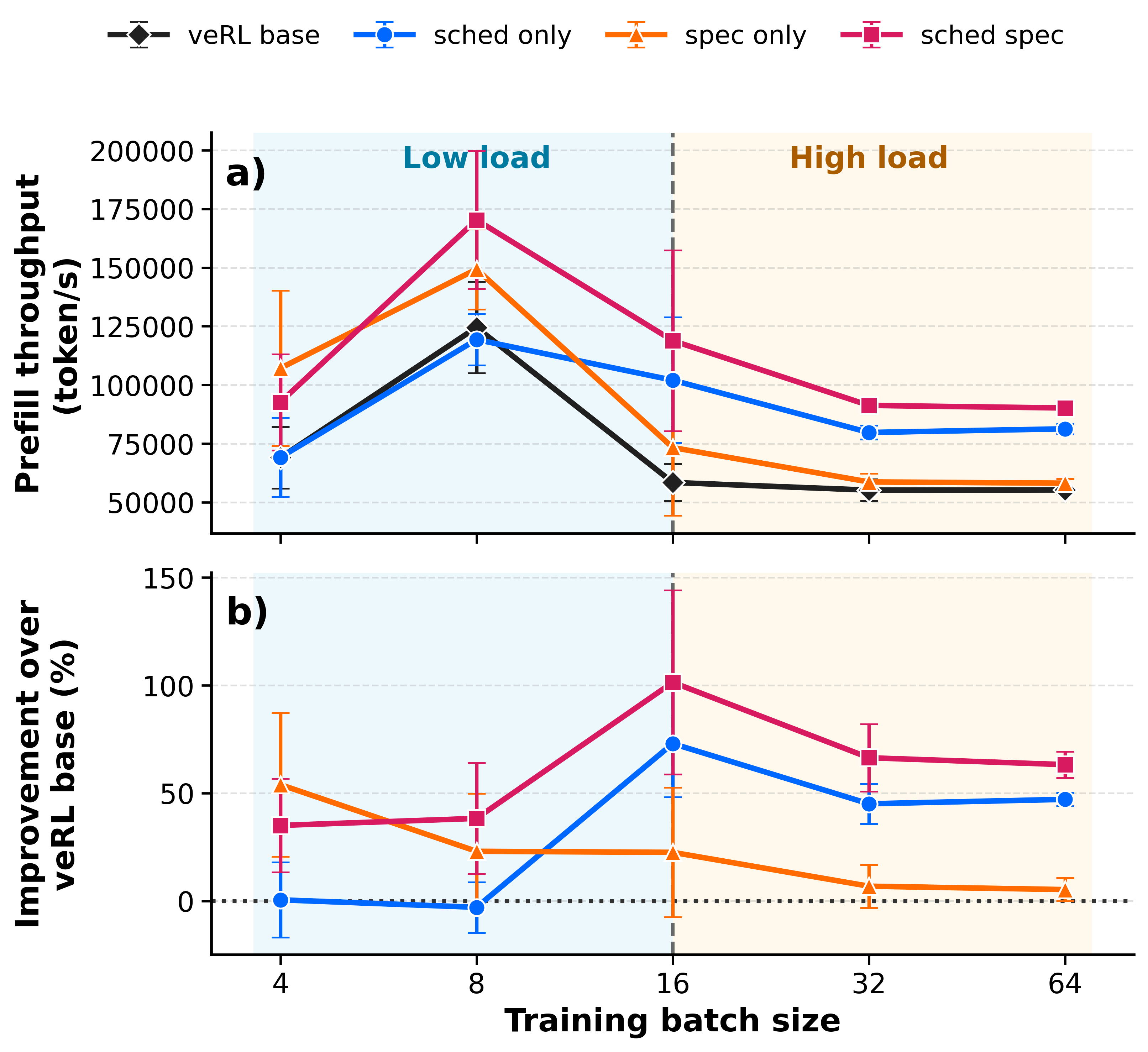}
    \caption{The top panel reports the absolute prefill throughput, while the bottom panel shows the relative improvement over \texttt{veRL\_base}. 
    Each point represents the mean over five RL steps, with error bars indicating the standard deviation. 
    Cache-aware scheduling provides increasingly larger gains as the workload grows, while the combined \texttt{sched\_spec} configuration achieves consistently high prefill throughput across workload regimes. 
    The vertical dashed line marks the boundary between the low- and high-load regimes.}
    \label{fig:prefillTpt}
\end{figure}

\begin{figure}[t]
    \centering
    \includegraphics[width=1.0\linewidth]{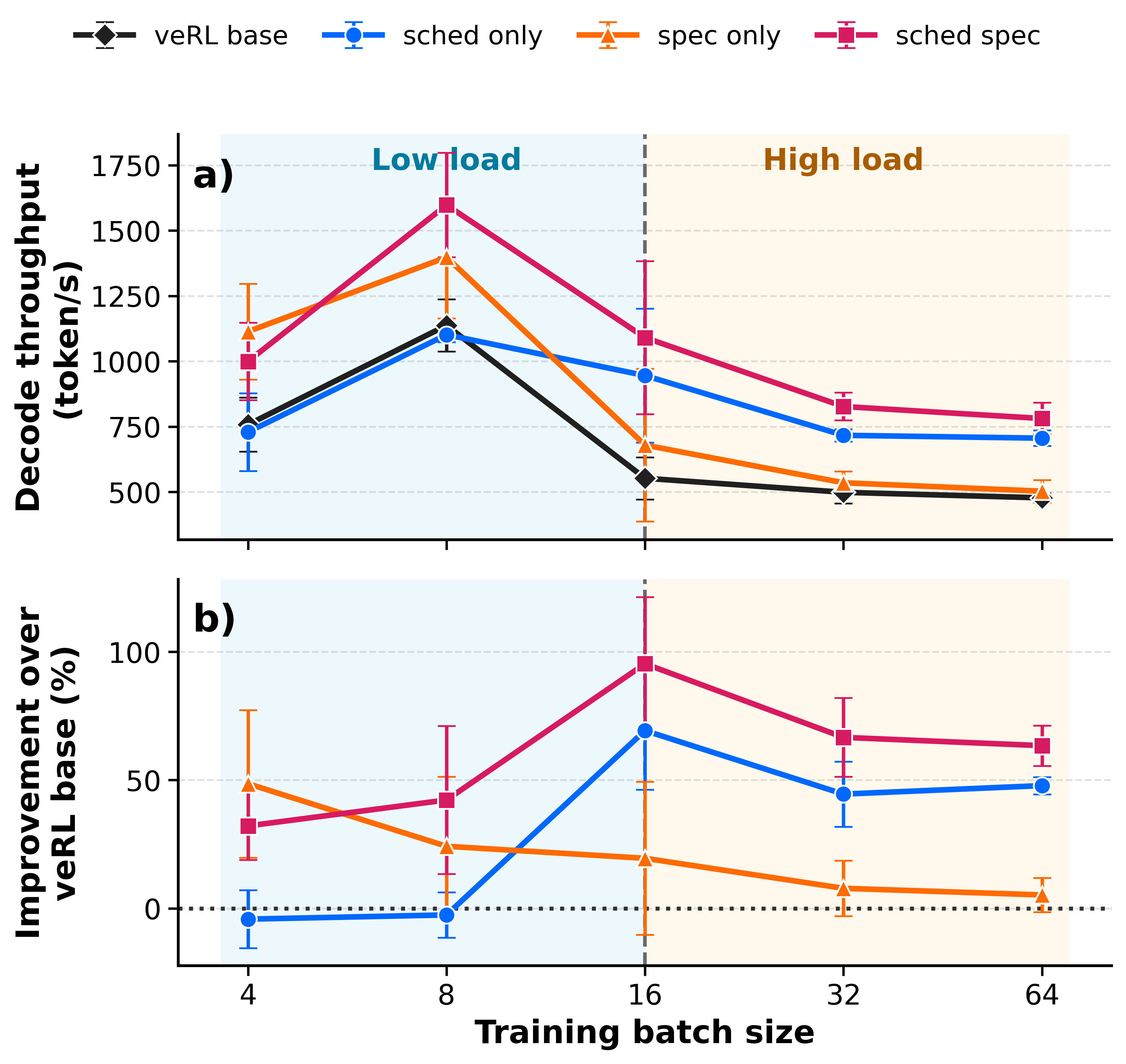}
    \caption{Decode throughput across different training batch sizes. 
    The top panel reports the absolute decode throughput, while the bottom panel shows the relative improvement over \texttt{veRL\_base}. 
    Each point represents the mean over five RL steps, with error bars indicating the standard deviation. 
    SuffixDecoding is most effective under low load, whereas cache-aware scheduling becomes increasingly beneficial under high load. 
    Combining the two mechanisms enables \texttt{sched\_spec} to maintain robust decode-throughput improvements across the full workload range. 
    The vertical dashed line marks the boundary between the low- and high-load regimes.}
    \label{fig:decodeTpt}
\end{figure}

\texttt{Sched\_only} becomes increasingly effective as workload grows. When the batch size increases 
from 4 to 64, relative improvement of in prefill throughput 
(Figure~\ref{fig:prefillTpt}) increases from 0\% to 47\%, while its improvement in decode throughput 
(Figure~\ref{fig:decodeTpt}) increases from -4\% to 47\%. This indicates that cache-aware scheduling 
is more effective under heavier load, especially when the batch size is no smaller than 16.

In contrast, \texttt{spec\_only} shows the opposite trend. Its relative improvement in prefill throughput 
decreases from 53\% to 5\%, and its improvement in decode throughput decreases from 48\% to 5\% as the batch size 
increases from 4 to 64. This suggests that SuffixDecoding is most effective under lower load, namely when the 
batch size is smaller than 16. 

\begin{figure}[t]
    \centering
    \includegraphics[width=1.0\linewidth]{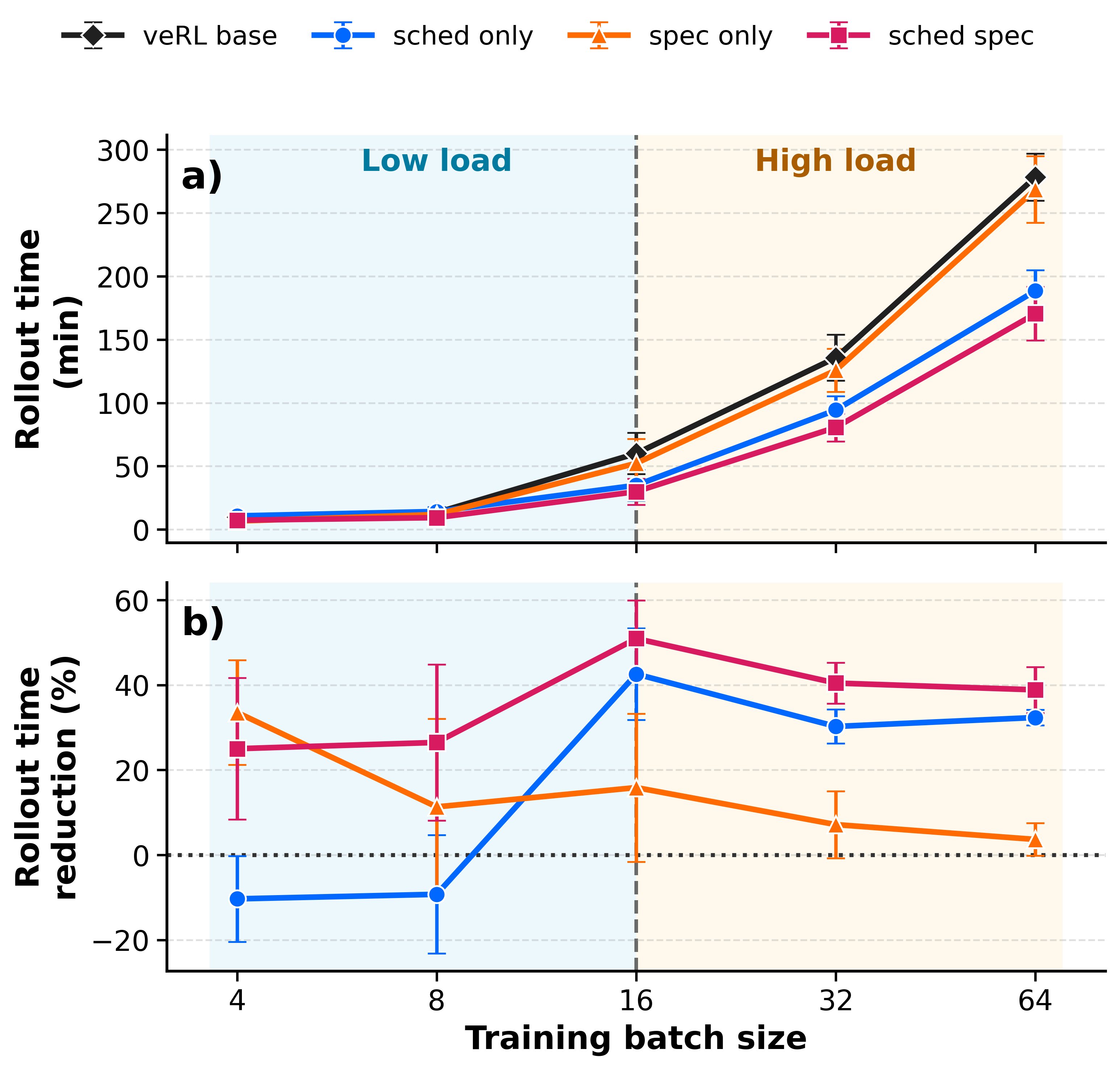}
    \caption{End-to-end rollout time across different training batch sizes. 
    The top panel reports the absolute rollout time, while the bottom panel shows the relative time reduction compared with \texttt{veRL\_base}, where a larger value indicates a greater speedup. 
    Each point represents the mean over five RL steps, with error bars indicating the standard deviation. 
    SuffixDecoding primarily reduces rollout time under low load, while cache-aware scheduling provides larger benefits as the workload increases. 
    Their combination achieves consistent end-to-end acceleration across all evaluated batch sizes. 
    The vertical dashed line marks the boundary between the low- and high-load regimes.}
    \label{fig:rolloutTime}
\end{figure}

By combining the two techniques, \texttt{sched\_spec} achieves robust throughput improvements across 
the full workload range. Its prefill throughput improvement increases from 35\% to 63\%, and its decode 
throughput improvement increases from 32\% to 63\% as the batch size grows from 4 to 64. A similar complementary 
effect is also observed in end-to-end rollout time, as shown in Figure~\ref{fig:rolloutTime}.

We further investigate cache hit rate and acceptance length, which are closely related to cache-aware 
scheduling and SuffixDecoding, respectively. 

\subsection{Cache-Aware Scheduling}

\begin{figure}[t]
    \centering
    \includegraphics[width=1.0\linewidth]{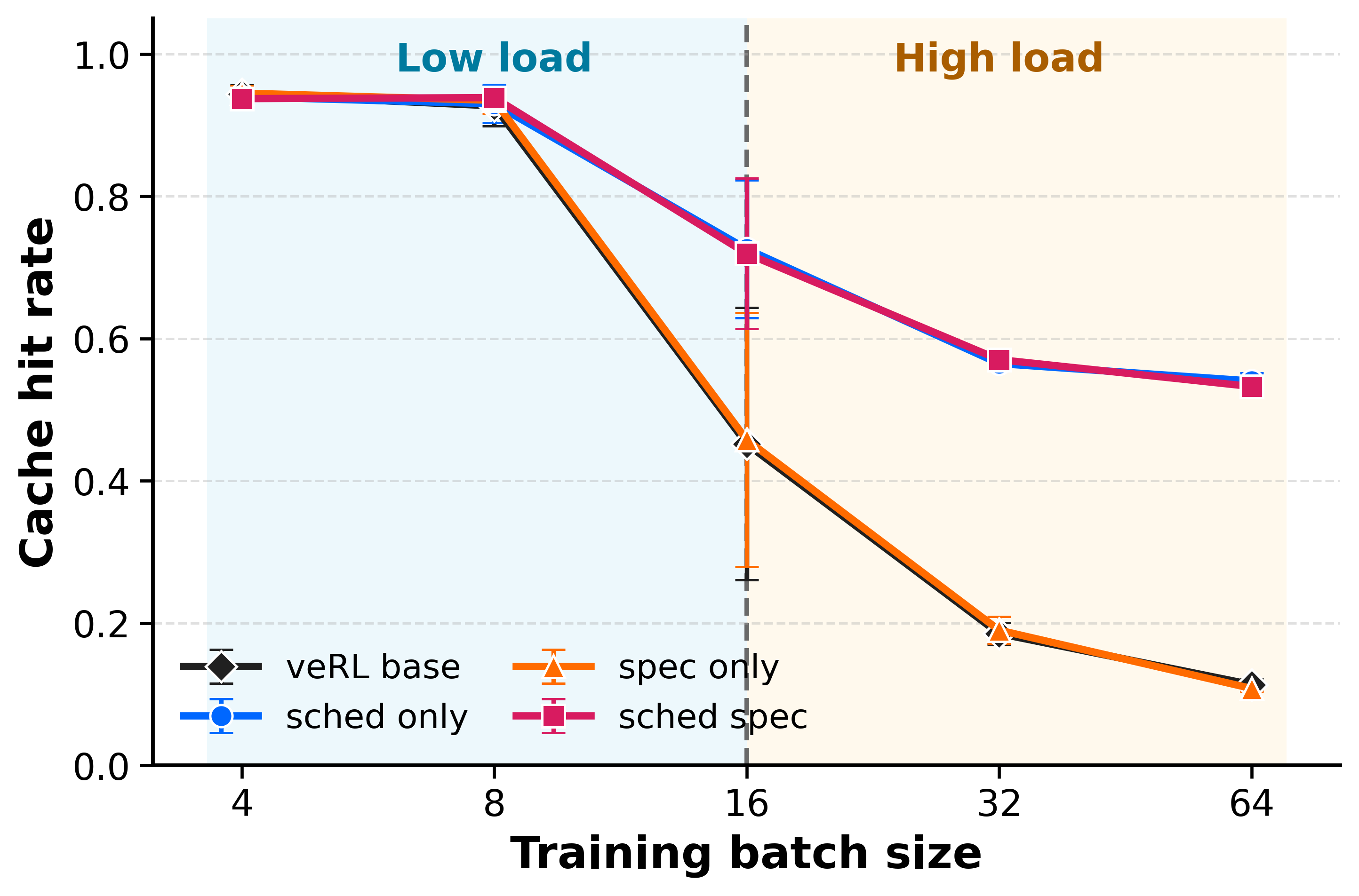}
    \caption{KV-cache hit rate across different training batch sizes.
    Each point reports the mean over five RL steps, with error bars indicating the standard deviation.
    Under low load, all configurations achieve similarly high cache hit rates.
    As the workload increases, the cache hit rate of \texttt{veRL\_base} and \texttt{spec\_only} drops substantially, 
    whereas configurations with cache-aware scheduling, \texttt{sched\_only} and \texttt{sched\_spec}, preserve 
    significantly higher cache locality. The vertical dashed line indicates the workload-regime boundary used in our analysis.}
    \label{fig:cacheHit}
\end{figure}

As shown in Figure~\ref{fig:cacheHit}, when the load is low, 
namely when the batch size is smaller than 16, all experiment settings achieve a very high cache hit rate, 
reaching 0.94 and 0.93 for batch sizes 4 and 8, respectively.

As the load increases, the cache hit rates of \texttt{veRL\_base} and \texttt{spec\_only} drop significantly, 
decreasing from around 0.9 at low load to 0.1 when the batch size reaches 64. This is because cache-aware 
scheduling is disabled in these two settings, and the default scheduling policy relies on least-request scheduling 
with sticky sessions. Least-request scheduling is unaware of KV-cache locality. Sticky sessions provide implicit 
cache awareness by routing requests from the same session to the same rollout instance, but the corresponding 
KV cache can still be evicted during long environment interactions when it is not revisited for a long time.

In contrast, with cache-aware scheduling enabled, both \texttt{sched\_only} and \texttt{sched\_spec} maintain a 
relatively high cache hit rate even under high load, achieving around 0.5 when the batch size is 64.

\begin{table}[t]
\centering
\caption{Effect of waiting factor on rollout presented with relative difference (\%) betweeen disabling \(S_\text{waiting}\) and enabling \(S_\text{waiting}\)}
\label{tab:waitingFactorAblation}
\begin{tabular}{lcc}
\toprule
\textbf{batch size} & \textbf{32} & \textbf{64} \\
\midrule
Prefill tokens & -11.7 & -44.5 \\
Decode tokens & -8.6 & -41.3 \\
Assistant turns & -9.4 & -38.9 \\
Rollout end-to-end time & -12.0 & -46.9 \\
\bottomrule
\end{tabular}
\end{table}

We perform an ablation study to evaluate the effect of the four scheduling factors introduced in 
Section~\ref{sec:cacheAwareScheduling}. We first examine the role of the waiting factor, 
\(\text{scale}_{\text{waiting}}\), by setting it to either zero or a non-zero value while keeping all 
other factors unchanged, as shown in Table~\ref{tab:waitingFactorAblation}.

As the rollout load increases, some queued requests may wait for a long time before being scheduled. 
If a request is not dispatched before the corresponding sandbox reaches its TTL, the sandbox shuts down and 
the agent trajectory is terminated prematurely. This leads to incomplete rollouts, which is reflected in 
the overall reduction of prefill tokens, decode tokens, and end-to-end time. As the load becomes heavier, 
i.e., when the batch size increases from 32 to 64, this rollout degradation becomes more severe.

\begin{figure}[!t]
    \centering
    \includegraphics[width=1.0\linewidth]{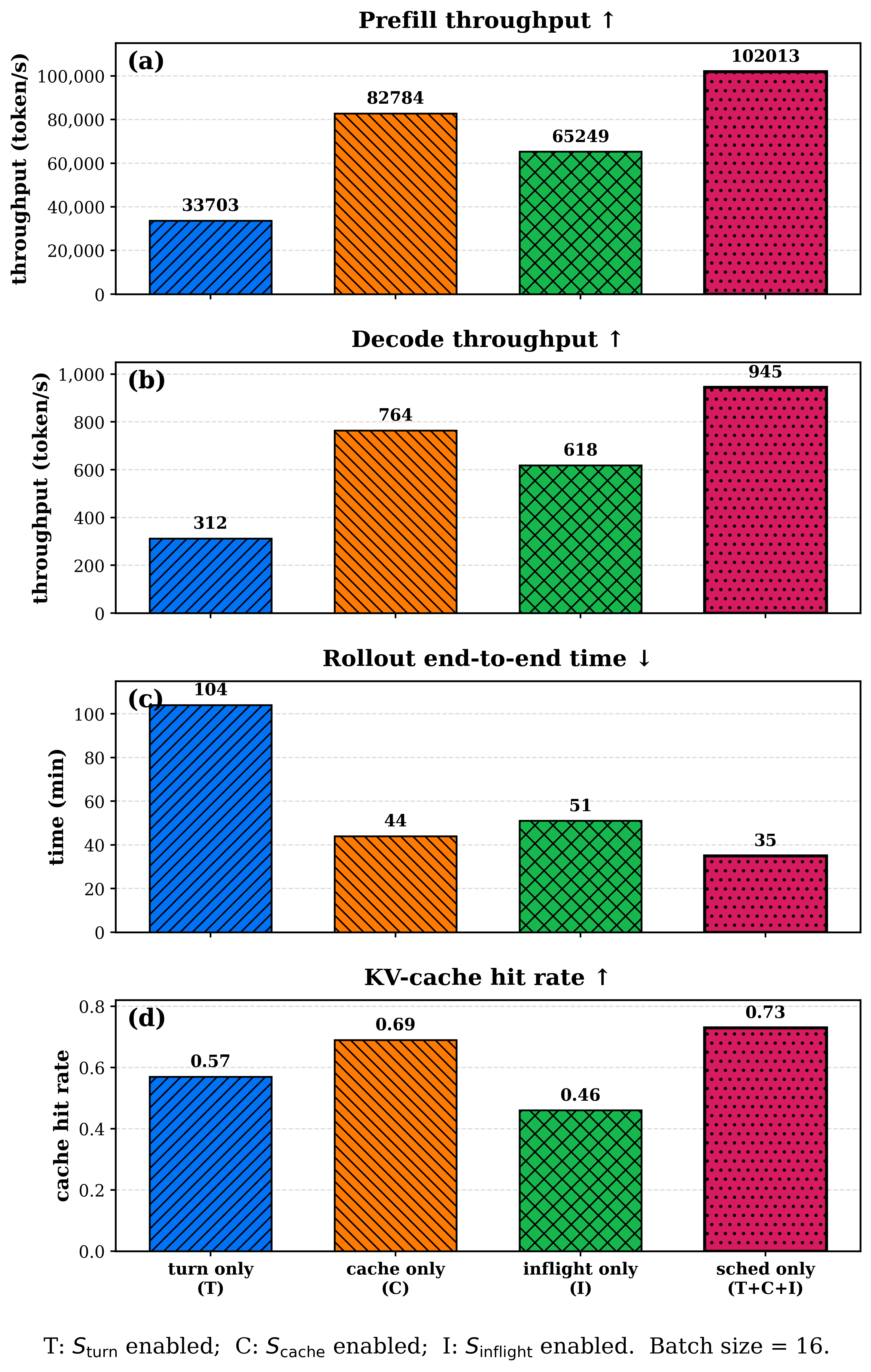}
    \caption{Ablation study of cache-aware scheduling at a training batch size of 16.
    The four panels report prefill throughput, decode throughput, rollout end-to-end time, and KV-cache hit rate, respectively.
    Among the single-factor variants, \texttt{cache\_only} achieves the best overall performance, highlighting the importance of KV-cache locality.
    The full \texttt{sched\_only} configuration, which jointly considers estimated trajectory length, cache locality, and replica load, achieves the highest prefill and decode throughput, the lowest rollout time, and the highest cache hit rate.
    These results show that the three scheduling factors are complementary.}
    \label{fig:cacheAwareAblation}
\end{figure}

For the ablation study of \(S_\text{turn}\), \(S_\text{cache}\), and \(S_\text{inflight}\), 
we keep \(S_\text{waiting}\) enabled to preserve the integrity of rollout trajectories, 
as shown in Figure~\ref{fig:cacheAwareAblation}. Among the three ablation settings, 
namely \texttt{turn\_only}, \texttt{cache\_only}, and \texttt{inflight\_only}, 
\texttt{cache\_only} achieves the best performance, with the highest throughput and cache hit rate. 
This highlights the importance of cache locality in rollout scheduling.

Notably, none of the ablation settings outperforms \texttt{sched\_only}, which corresponds to the full 
scheduling configuration. This indicates that the three scheduling components are complementary and 
jointly contribute to the overall performance.

\subsection{SuffixDecoding}

\begin{figure}[t]
    \centering
    \includegraphics[width=1.0\linewidth]{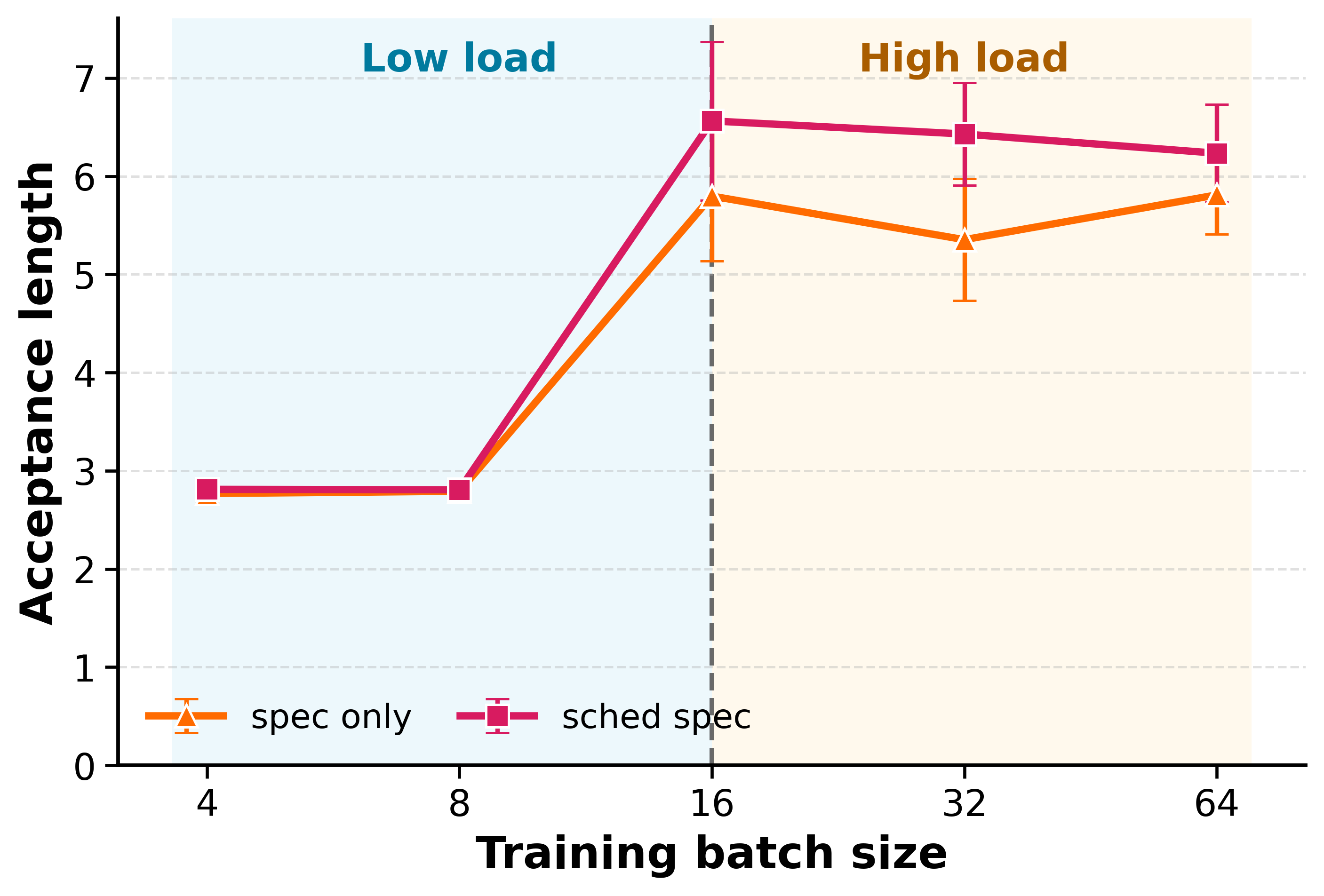}
    \caption{Average acceptance length of SuffixDecoding across different training batch sizes.
    Each point reports the mean over five RL steps, with error bars indicating the standard deviation.
    Both \texttt{spec\_only} and \texttt{sched\_spec} achieve acceptance lengths substantially greater 
    than one, demonstrating that recurring token patterns in agentic rollouts provide useful suffix-based drafts.
    Although the acceptance length remains high under heavier workloads, this does not necessarily translate into 
    higher decode throughput because verification overhead becomes more significant as GPU utilization increases.
    Configurations without SuffixDecoding are omitted. The vertical dashed line indicates the workload-regime 
    boundary used in our analysis.}
    \label{fig:accLen}
\end{figure}

As shown in Figure~\ref{fig:accLen}, when SuffixDecoding is enabled, i.e., in \texttt{spec\_only} and 
\texttt{sched\_spec}, the acceptance length reaches around 2.8 when the batch size is smaller than 16, 
and around 6.0 when the batch size is no smaller than 16. This demonstrates that SuffixDecoding can 
effectively generate useful draft tokens for SWE tasks. We further observe from the rollout logs that 
the model often generates recurring code snippets, which provide favorable matching opportunities for SuffixDecoding.

However, the high acceptance length does not always translate into higher throughput. A key observation of 
speculative decoding is that its benefit depends on the batch size. When the batch size exceeds a certain 
threshold, speculative decoding may no longer yield performance gains, and disabling it can result in better 
throughput. This is because speculative decoding is most beneficial when normal decoding underutilizes the GPU. 
At small batch sizes, verifying multiple draft tokens in one forward pass increases parallelism and improves 
hardware utilization. As the batch size grows, normal batched decoding already saturates GPU resources, leaving 
limited room for additional speculative speedup. Meanwhile, speculative decoding still incurs extra overhead from 
draft generation, verification, and tree-mask management. When this overhead outweighs the reduction in decoding steps, 
disabling speculative decoding becomes more efficient.

\begin{figure}[t]
    \centering
    \includegraphics[width=1.0\linewidth]{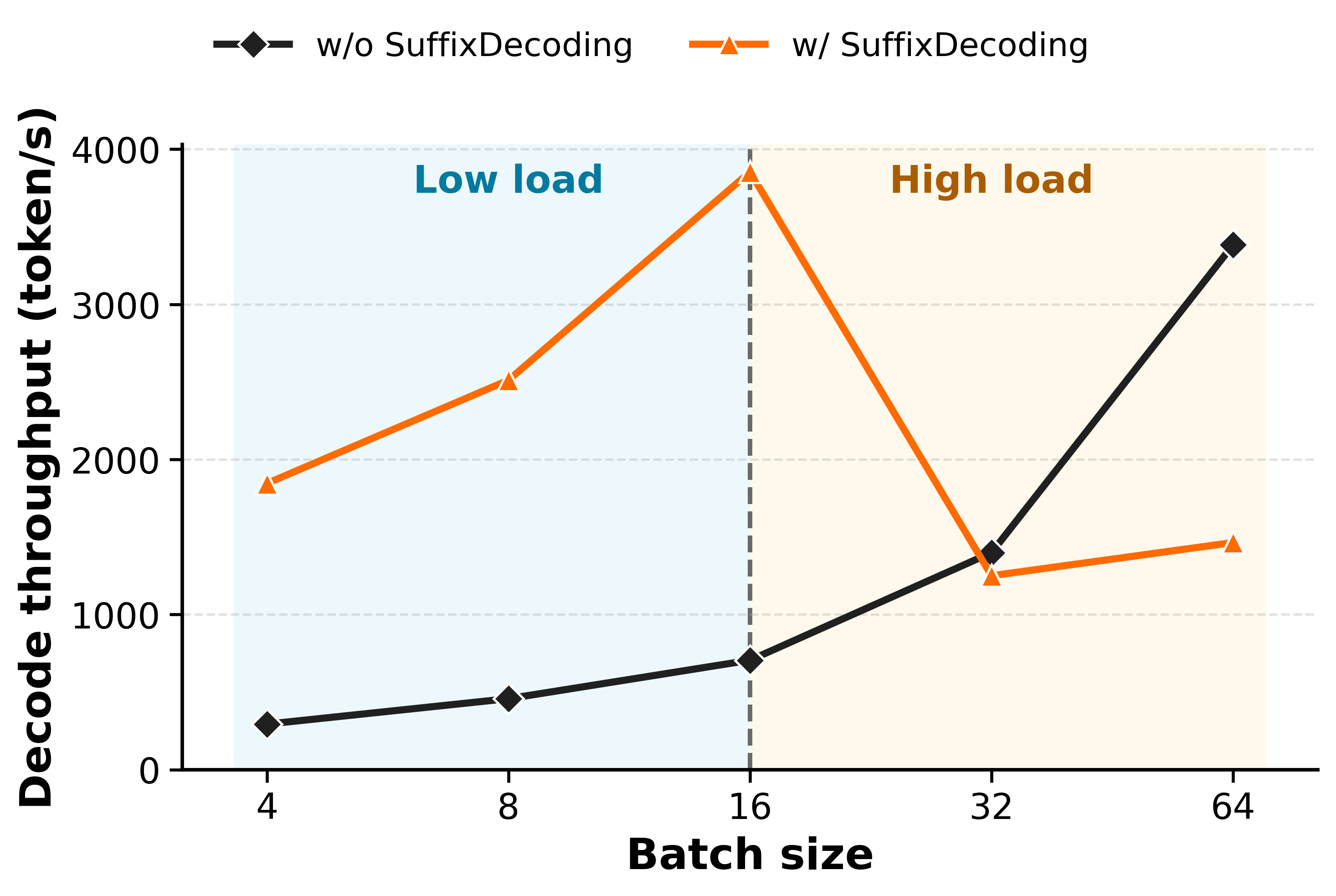}
    \caption{Effect of batch size on decode throughput with and without
    SuffixDecoding using Qwen3-32B. The input and output lengths are fixed
    at 8,192 and 512 tokens, respectively. SuffixDecoding substantially
    improves decode throughput at small batch sizes, where conventional
    batched decoding underutilizes the GPU. As the batch size increases,
    the benefit diminishes and eventually becomes negative because normal
    decoding already achieves high GPU utilization while speculative
    verification introduces additional overhead. The vertical dashed line
    marks the workload-regime boundary used in our analysis.}
    \label{fig:batchOnDecTpt}
\end{figure}

We observe the same phenomenon for SuffixDecoding. As shown in Figure~\ref{fig:batchOnDecTpt}, once the batch 
size exceeds 16, SuffixDecoding begins to underperform the non-SuffixDecoding baseline in decode throughput. 
This result indicates that SuffixDecoding should be enabled under low load but disabled under high load.

We conduct an ablation study to analyze the sources of the high acceptance length, as shown in 
Table~\ref{tab:accAblation}. The results indicate that SuffixDecoding benefits from both cross-prompt 
similarity and intra-prompt rollout similarity. Different prompts may generate similar code fragments, 
while multiple rollouts of the same prompt often share similar local structures. Both types of recurring 
patterns provide useful suffix matches for draft generation and contribute to the high acceptance length.

\begin{table}[t]
\centering
\caption{Acceptance length at the 1st step of training under different batch size and rollout\_n.}
\label{tab:accAblation}
\begin{tabular}{cccc}
\toprule
{\textbf{batch size}} 
& \multicolumn{3}{c}{\textbf{rollout\_n}} \\
\cmidrule(lr){2-4}
& \textbf{4} & \textbf{8} \\
\midrule
8  & 2.7 & 2.7 \\
16 & 2.6 & 5.5 \\
\bottomrule
\end{tabular}
\end{table}

\section{Related Work}

\subsection{RL Training Systems}

Recent RL training systems improve the efficiency of LLM post-training from different perspectives. 
Stage-overlap systems, such as RLHFuse~\cite{zhong2025optimizingrlhftraininglarge}, reduce pipeline 
bubbles by fusing rollout, reward computation, and training-related stages. Other systems improve 
resource utilization through asynchronous execution or relaxed synchronization~\cite{sheng2025laminarscalableasynchronousrl, fu2026areallargescaleasynchronousreinforcement, zhong2025streamrlscalableheterogeneouselastic, hu2026dorascalableasynchronousreinforcement}. 
These approaches can reduce end-to-end training latency, but they either overlap non-dominant stages 
with rollout or relax strict on-policy semantics. In contrast, \systemname{} focuses on accelerating 
synchronous rollout itself, without changing the underlying RL algorithm or relaxing the synchronization boundary.

\subsection{Rollout Scheduling for Synchronous RL}

Several recent systems address the long-tail inefficiency of synchronous rollout through scheduling. 
RollPacker~\cite{gao2025rollpackermitigatinglongtailrollouts} proposes tail batching, which consolidates 
prompts likely to produce long responses into a small number of long rounds while keeping most rounds short and balanced. 
Seer~\cite{qin2026seeronlinecontextlearning} decomposes prompt groups into finer-grained chunks and uses 
context-aware scheduling to reduce load imbalance and KV-cache pressure. These systems show that scheduling is 
crucial for efficient synchronous RL rollout. \systemname{} is complementary to them: instead of focusing 
only on long-tail scheduling, it treats rollout optimization as workload-dependent and combines cache-aware 
scheduling with decoding-level acceleration.

\subsection{Speculative Decoding for RL Rollout}

Speculative decoding accelerates autoregressive generation by drafting multiple tokens and verifying them 
with the target model in parallel. For RL rollout, recent systems adapt speculative decoding to address 
model drift, long-tail latency, and distributional correctness. TLT~\cite{hu2026taminglongtailefficientreasoning} 
maintains an adaptive model-based drafter using idle GPUs during the long-tail phase. 
DAS~\cite{shao2025beatlongtaildistributionaware} constructs a distribution-aware suffix-tree drafter from recent 
rollouts and allocates more speculative budget to long-tail trajectories. These methods demonstrate the potential 
of speculative decoding in RL training, but speculative decoding is not uniformly beneficial across all workload regimes. 
\systemname{} therefore applies model-free SuffixDecoding under low load, where GPU resources are underutilized, 
and disables it under high load, where speculation overhead can outweigh its benefits.

\subsection{LLM Serving and Cache-Aware Scheduling}

Efficient LLM serving systems improve throughput through continuous batching, KV-cache management, prefix caching, 
and request scheduling. These techniques are largely orthogonal to RL training, but they become increasingly 
important for long-context agentic rollouts, where KV-cache recomputation and memory pressure dominate high-load 
performance. \systemname{} brings cache-aware scheduling into synchronous agentic RL rollout by considering cache 
locality, trajectory progress, and server load when assigning requests to rollout replicas. Unlike general-purpose 
serving systems, \systemname{} exploits RL-specific workload structure and adapts its optimization strategy 
according to runtime load.
\section{Conclusion}

This paper argues that workload should be treated as a first-class factor in optimizing 
synchronous agentic RL rollouts. Long-horizon agent trajectories create distinct bottlenecks under 
different runtime loads: low-load rollouts suffer from insufficient GPU utilization, while high-load 
rollouts are dominated by KV-cache pressure, redundant recomputation, and load imbalance. 
We propose \systemname{}, a workload-aware rollout system that addresses these bottlenecks by jointly 
optimizing decoding and scheduling.

\systemname{} applies model-free SuffixDecoding under low load to reuse historical suffix patterns 
as speculative drafts without introducing draft-model GPU contention. Under high load, \systemname{} 
shifts the optimization focus to cache-aware scheduling, placing requests across rollout replicas based on 
cache locality, trajectory progress, and server load. This simple design preserves synchronous RL semantics 
and remains compatible with existing RL training pipelines. Experiments on long-context agentic workloads 
show that \systemname{} improves rollout throughput by $1.4\times$ under low load and up to $1.6\times$ under high load 
over the veRL baseline. These results show that workload-aware rollout optimization provides a practical path 
toward scalable synchronous agentic RL training.

\bibliography{ref}

\end{document}